\pdfoutput=1
\documentclass[11pt]{article}
\usepackage[final]{ACL2023}
\usepackage{multirow}
\usepackage{fontawesome5}
\usepackage{graphicx}
\usepackage{booktabs} %
\usepackage[inline]{enumitem}
\usepackage{mdframed}
\usepackage{makecell}
\usepackage{enumitem}
\setlist[itemize]{noitemsep, nolistsep}
\usepackage{listings}
\usepackage{tcolorbox}
\tcbuselibrary{listings,skins}
\usepackage{times}
\usepackage{latexsym}
\usepackage[T1]{fontenc}
\usepackage{lipsum}
\usepackage[utf8]{inputenc}
\usepackage{booktabs}
\usepackage{microtype}
\usepackage{inconsolata}
\usepackage{verbatimbox}
\usepackage{soul}
\usepackage{algorithm,algpseudocode}
\usepackage{multicol}
\usepackage{amsmath}
\usepackage{enumitem}
\newcommand{\header}[1]{\vspace*{1mm}\noindent\textbf{#1}.}
\title{SOLID: Self-seeding and Multi-intent Self-instructing LLMs for Generating Intent-aware Information-Seeking Dialogs}

\author{Arian Askari\textsuperscript{1}\thanks{* Equal contribution (shared co-authorship)} , Roxana Petcu\textsuperscript{2}\footnotemark[1] , Chuan Meng\textsuperscript{2} , Mohammad Aliannejadi\textsuperscript{2}  \\ \textbf{Amin Abolghasemi\textsuperscript{1}} , \textbf{Evangelos Kanoulas\textsuperscript{2}} , \textbf{Suzan Verberne\textsuperscript{1}} \\
\textsuperscript{1}Leiden University \{a.askari, m.a.abolghasemi, s.verberne\}@liacs.leidenuniv.nl \\
\textsuperscript{2}University of Amsterdam \{r.m.petcu, c.meng, m.aliannejadi, e.kanoulas\}@uva.nl\\
}

\begin{document}
\maketitle

\begin{abstract}
Intent prediction in information-seeking dialogs is challenging and requires a substantial amount of data with human-labeled intents for effective model training.
While Large Language Models (LLMs) have demonstrated effectiveness in generating synthetic data, existing methods typically rely on human feedback and are tailored to structured, task-oriented intents.
In this paper, we leverage LLMs for zero-shot generation of large-scale, open-domain, intent-aware information-seeking dialogs to serve as training data for intent prediction models.
We introduce SOLID, a method that generates dialogs turn by turn using novel \textit{self-seeding} and \textit{multi-intent self-instructing} strategies.
Additionally, we propose SOLID-RL, a finetuned version that generates an entire dialog in one step using data created with SOLID.
SOLID and SOLID-RL are each used to generate over 300k intent-aware dialogs, significantly surpassing the size of existing datasets.
Experiments show that intent prediction models trained on sampled dialogs generated by SOLID and SOLID-RL outperform those trained solely on human-generated dialogs.
Our findings demonstrate the potential of LLMs to expand training datasets, as they provide valuable resources for conversational agents across multiple tasks. Our self-seeding and self-instructing approaches are adaptable to various conversational data types and languages with minimal modifications.

\end{abstract}
\section{Introduction} 
\begin{figure}
    \centering
    \scalebox{.61}{\includegraphics[]{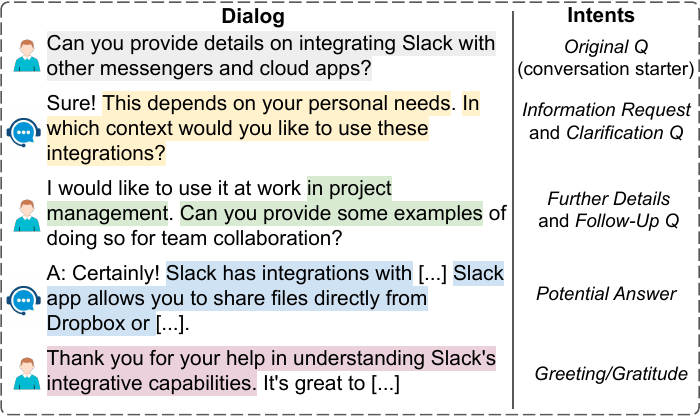}} 
    \caption{Example dialog with sequence of intents.}
    \label{fig:intent_example}
\end{figure}

Understanding user intent in information-seeking dialogs is essential for conversational agents to fulfil the user's information need. Specifically, identifying the underlying intent behind each conversational turn enables the dialog system to respond appropriately. For instance, if a user exhibits signs of confusion regarding the previous dialog turn, the system can reiterate its response with a more clear phrasing. A thorough definition of intents can be found in Appendix \ref{app:intents}. The turn-level intent identification task is called Intent Prediction\footnote{Intent prediction and intent classification are used interchangeably in this work.} (IP) \cite{zamani2023conversational}.
Leveraging predicted intents has been shown to benefit various Information Retrieval (IR) and Natural Language Processing (NLP) tasks, e.g., response selection~\citep{IART2020,kumar2019practical}, answer selection and user satisfaction estimation~\citep{deng2023intent,deng2022DialogActs}.
Figure~\ref{fig:intent_example} illustrates an example of intents in an information-seeking conversation extracted from the MSDialog-intent dataset \citep{qu2018MSDialog}. These are the intents referenced throughout this study. The full taxonomy of MSDialog intents can be found in Table \ref{table:intent_taxonomy} in Appendix~\ref{sec:dialog_generation}. Notably, intents are defined at the utterance level, with each utterance being associated with one or potentially multiple intents. Using MSDialog intents ensures compatibility with existing datasets, however, our methodology is inherently flexible, as it can accommodate finer-grained taxonomies or additional annotations.

\par
\header{Motivation}
Intent Prediction (IP) is a challenging task~\citep{qu2019UserIntentPred}. 
Recent work~\citep{zhao2023chatgpt} has shown that prompting ChatGPT to handle IP tasks in a few-shot manner reaches limited performance, with ChatGPT performing worse than supervised IP methods.
Supervised methods, however, require a large number of human-annotated information-seeking dialogs to achieve satisfactory performance, and often fail to generalize across different datasets.
For example, \citet{InforSeek_User_Intent_Pred} show that their IP model trained on MSDialog-intent achieves a precision of 69\% compared to only 47\% on the Ubuntu dialog corpus%
. This large gap in effectiveness might be due to the small size of existing datasets, which limits the ability to generalize.
Furthermore, manually annotating intents is resource-intensive, leading to limited training data for IP models~\citep{InforSeek_User_Intent_Pred,penha2019introducing}. %
Using LLMs directly for intent classification is not a reliable alternative, since it has been shown that LLMs have limited effectiveness in accurately labeling intents~\cite{weiwei2023diqad,unlocking2023,zhao2023chatgpt}. 

We employ LLMs for large-scale synthetic data augmentation%
, where each dialog utterance has one or more associated intents. Our proposed method generates each dialog by following a turn-by-turn sequence of real-world intents extracted from MSDialog-intent \citep{qu2018MSDialog}. Leveraging these intents from existing human-labeled dialogs ensures that our synthetic data aligns and possesses the same level of diversity as real-world data. While MSDialog-intent is built on question answering (QA) interactions between information seekers and product specialists on Microsoft products, it is presented as a dialog dataset. Each dialog is entity-centric, i.e., it is constructed around a specific entity as the object of discussion. %
\par
\begin{figure}[]
    \centering
    \scalebox{.45}{\includegraphics[]{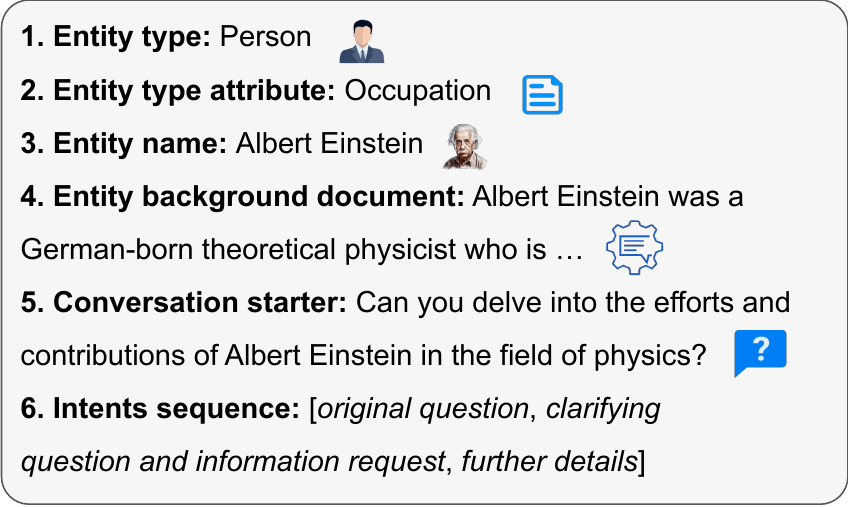}} %
    \caption{Example of a seed generated through self-seeding via SOLID.}
    \label{fig:self-seed}
\end{figure}

\header{Challenges}
We identify two main challenges in using LLMs for generating intent-aware information-seeking dialogs:  
\begin{enumerate*}[label=(\roman*)]
    \item Information-seeking dialogs are typically centred around a topic or entity that serves as the foundation of the conversation. In dialog generation, this information is known as the seed of the dialog, providing the starting point and context of the conversation~\citep{kulkarni2024synthdst}.
    An example of a dialog seed is shown in Figure \ref{fig:self-seed}. While the common practice is to leverage seed information from external resources (e.g., Wikipedia)~\cite{ding2023enhancing}, recent work has shown that feeding such seeds to an LLM often results in low-quality generations, as LLMs tend to underperform when prompted with unfamiliar information.
    \item Our preliminary results show that prompting an open-source LLM to generate an entire dialog yields low performance, often producing too many or too few utterances that do not follow the intent sequence. However, generating utterances one by one improves generation quality.
\end{enumerate*}
\par
\header{SOLID}
To address these challenges, we propose SOLID: a \textbf{S}elf-seeding and multi-intent self-instructing method f\textbf{o}r generating \textbf{L}arge-scale \textbf{I}ntent-aware Information-Seeking \textbf{D}ialogs.
SOLID has novel \textit{self-seeding} and \textit{multi-intent self-instructing} features:
\begin{enumerate*}[label=(\roman*)]
    \item \textit{self-seeding} involves querying the LLM to generate a seed, which is then used in a subsequent step to prompt the same LLM to produce a dialog.
    \item \textit{multi-intent self-instructing} allows our framework to generate utterances conditioned on multiple intents simultaneously. %
\end{enumerate*}
\par
\begin{figure*}[t]
    \centering
    \includegraphics[]{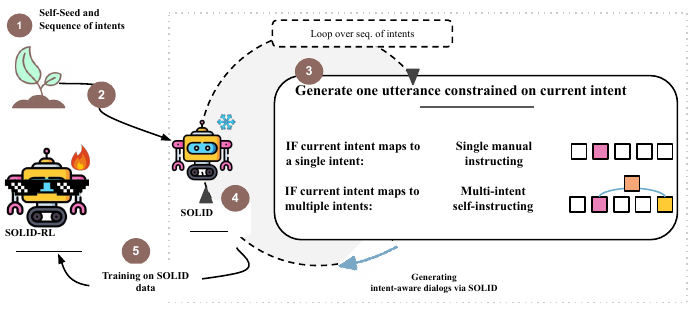} 
    \caption{Illustrating SOLID and SOLID-RL: Starting with a self-generated seed and a sequence of real-world intents, SOLID produces utterances sequentially, with each utterance conditioned on one or more intents. During phase 3, each utterance falls into one of two categories: it either corresponds to a single intent, based on a definition from taxonomy \ref{table:intent_taxonomy}, or combines multiple intent definitions. In the case of multiple intents, SOLID employs multi-intent self-instruction to merge these into a single cohesive instruction.} 
    \label{fig:solid_rl}
\end{figure*}

\header{SOLID-RL}
While SOLID creates high-quality, intent-aware conversations by generating dialogs turn by turn, it incurs significant computational costs. To enhance efficiency, we propose SOLID-RL, which enables the generation of the full dialog in a single pass. Although prompting an LLM to generate an entire dialog reduces quality, similar to how solving complex tasks without step-by-step breakdowns results in less coherent reasoning \citep{wei2022CoT}, SOLID-RL mitigates this problem by fine-tuning the LLM on the high-quality dialogs produced by SOLID. This method preserves dialog quality while significantly reducing computational overhead. The framework is illustrated in Figure \ref{fig:solid_rl}.
\par
\header{Contributions} 
Our contributions are as follows: 
\begin{itemize}[leftmargin=*]
    \item SOLID: We propose a novel \textbf{S}elf-seeding and multi-intent self-instructing method f\textbf{o}r generating \textbf{L}arge-scale \textbf{I}ntent-aware Information-Seeking \textbf{D}ialogues.
    \item SOLID-RL: We propose a fine-tuned method for efficiently generating entire dialogs using SOLID-generated synthetic dialogs.
    \item We provide two large-scale synthetic intent-aware information-seeking dialog datasets: SOLISpeak and SOLITurbo, each consisting of 316,697 dialogs. Our implementation is open-sourced as a Python library.\footnote{\href{https://anonymous.4open.science/r/solid\_repo/README.md}{https://anonymous.4open.science/r/solid\_repo/README.md}}
    \item We demonstrate that IP methods trained on a mix of limited available human data and a subset of SOLISpeak and SOLITurbo achieve significantly higher effectiveness than those trained solely on human-labeled datasets.
\end{itemize}
\section{Related work}
\header{Data augmentation with LLMs}
Data augmentation using LLMs has gained significant attention in recent research \cite{abolghasemi-2024-cause,askari2023expand, tunstall2023zephyr, abbasiantaeb2023let, liu2024ChatQA}. 
For example, ConvSDG is a framework that improves conversational search performance by leveraging synthetic training data generated at the dialog, session and query level ~\cite{mo2024ConvSDG}. Similarly, LAPS ~\cite{joko2024LAPS} introduces a data augmentation framework where LLMs guide a human worker into constructing personalized multi-session dialogs, enabling the creation of large-scale conversational datasets that reflect real-world user preferences. \citep{Lin2024GenRefine} further contributes with a zero-shot method to generate refined data for intent detection, expanding the versatility of generated data for downstream tasks. The Self-Instruct method ~\cite{wang2022selfinstruct} allows LLMs to augment their own prompt instructions~\cite{alpaca,vicuna2023,zheng2023lmsyschat1m}. %
Dialogic~\cite{gu2023dialogic} employs in-context learning to generate annotated dialogues in a controllable fashion, leading to a more structured data augmentation framework. Meanwhile, SynTOD~\cite{jin2024syntod} introduces an approach that simulates task-oriented dialogues using both state transition graphs and LLMs. However, none of these existing datasets have user intents associated, therefore they cannot be used for training IP models. %
Additionally, some studies use LLMs to augment training data for intent classification, but the utterances lack dialog context ~\cite{sahuetal2022data} or are limited to "yes" and "no" responses \cite{kim2022generating}, which differs from intent-aware conversation generation. %
\par
\header{Intent Prediction in information-seeking dialogs}
Utterance-level IP in dialog systems is a well-studied task aimed at identifying the intent at each utterance within a dialog~\citep{meng2023system,colombo2020guiding}.
MSdialog-Intent~\cite{qu2018AnalyzindMSDialog} and MANtIS~\cite{penha2019mantis} are the largest available datasets on intent-aware information-seeking dialogs (statistics are shown in Table \ref{tab:stat}). 
However, these datasets still remain relatively small, because of the high cost of human annotations. As stated by \cite{InforSeek_User_Intent_Pred} annotating utterance intents of 2,100 information-seeking conversations costs approximately 1,700 dollars using Amazon MTurk. 
This study focuses on open-domain information-seeking dialogs, excluding research on IP in other types of dialogs, such as chit-chat dialogs~\citep{ghazarian2022DialogAct,qu2023DialogActs}, dialogs with evidence from single or multi-knowledge sources~\cite{wu2023inscit}, and task-oriented dialogs~\cite{shah2018M2M, rastogi2020SGD, king2024IPTOD, arora2024intentdetectionagellms, ask2023closer}. 
\section{Intent-Aware Dialog Generation}
\subsection{SOLID}
\label{subsec:solid}
SOLID generates dialogs almost entirely based on its internal knowledge, without relying on external seeds or conversations, with the exception of using intent labels from MSDialog-intent. This ensures a self-contained generation process. 

\begin{figure*}[ht]
    \centering
    \scalebox{0.45}{\includegraphics[]{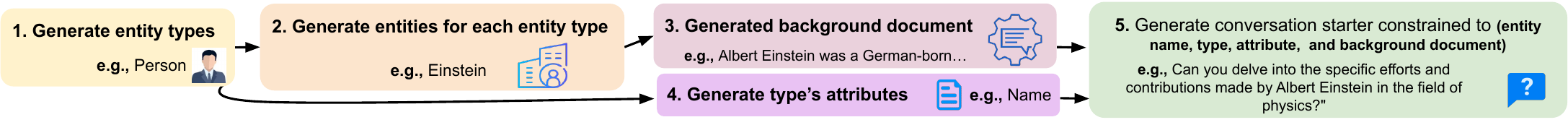}}
    \caption{The illustration of seed generation through self-seeding via SOLID.}
    \label{fig:self_seeding_process}
\end{figure*}

\header{Seed Generation}
SOLID is initiated by producing dialog seeds that serve as the starting point for dialog generation. Using the same model for both seed and dialog generation ensures that the LLM has internal knowledge of the entity and its attributes, enabling a more informed dialog generation process. %
Methods such as \citet{liu2024ExternalSeed} employ external seeds, relying on predefined content to generate dialogs using LLMs. While external seeds ensure factual background content, they do not guarantee that the LLM is familiar with specific information.
See Figure \ref{fig:self-seed} as an example of a seed that can be generated by SOLID. 
We take multiple steps to generate high-quality dialog seeds:
\begin{itemize}[nosep,leftmargin=*]
    \item \textit{Entity type and attribute generation:} We first prompt the LLM to generate 100 \textit{entity types} (e.g., `Person', 'Artist'), along with 10 \textit{entity attributes} associated with each entity type (e.g., `Occupation', 'Financial').    
    \item \textit{Entity name generation:} For each entity type, we ask the LLM to generate an \textit{entity name} (e.g., `Albert Einstein'). To encourage diversity in the data, we prompt the LLM to generate 100 names for each letter in the English alphabet across all entity types ($26 \times 100 \times 100$), resulting in 50K entity names after automatic filtering. More details on filtering entities in Appendix \ref{app:hallucinations}.
    \item \textit{Background document generation:} Inspired by \citet{yu2022generate}, we prompt the LLM to generate a \textit{background document} for each entity name (e.g., ``Albert Einstein was a ...'').
    \item \textit{Conversation starter:} Given the seed information generated so far, we prompt the LLM to create the first utterance, namely, the \textit{conversation starter} (e.g., ``Can you delve ...''). This is done by using the prompt shown in Figure \ref{fig:gen_conv_starter}.
    \item \textit{Intent sequence:} We randomly sample a \textit{sequence of intents} %
    from the existing MSDialog-intent dataset \citep{qu2018MSDialog}, which defines a set of intents for information-seeking conversations.  By performing sampling with replacement, we expect the distribution of intent sequences to closely match the original distribution of the dataset. This intent set allows us to capture the natural flow and complexity of real dialogs. The number of intents in the sampled sequence of intents indicates the length of the dialog to be generated. E.g., the number of intents in Figure \ref{fig:intent_example} is five, resulting in a dialog with five utterances.     
\end{itemize}
Figures~\ref{fig:self_seeding_process} and \ref{fig:solid_selfseed_detailed} illustrate the seed generation process in more detail. All generated entity types are illustrated in Appendix \ref{app:self_seeding}. Furthermore, all prompts are available in Appendix \ref{sec:prompts}.
\par
\header{Dialog Generation} Given a seed, we prompt the LLM to generate each subsequent utterance based on its associated intent. If the utterance is associated with a single intent, we extract the corresponding prompt instruction from Tables \ref{tab:agent_instructions} and \ref{tab:user_instructions} (details in Appendix~\ref{app:manual_instructions}). If the utterance is associated with multiple intents, i.e. any combination of two or more without repetition, we first retrieve the prompt instructions for each intent individually, and then query the LLM to create a customized prompt by merging the extracted instructions. \textit{Multi-intent self-instructing} is essential, as manually creating prompt instruction for $C_{12}^k$ as $k$ approaches larger values becomes unmanageable and impractical. Following Algorithm~\ref{algo:generate_dialog}, we generate each utterance and iteratively prompt the LLM with the seed, the next corresponding instruction, and the current dialog history. 

\header{Post-processing} In the final step, we post-process the generated dialogs and filter out the noisy ones. This involves three steps: (i) trimming to the last punctuation, as LLMs can reach their maximum token capacity without ending the last utterance, (ii) removing empty lines, and (iii) removing the dialog turn keywords, i.e., ``User'' or ``Agent.'' \footnote{Post-processing implementation is available at \href{https://anonymous.4open.science/r/solid\_repo/solid\_conversation/utils.py}{https://anonymous.4open.science/r/solid\_repo/solid\_conversation/utils.py}}

\begin{table}[]
    \centering
    \scalebox{0.60}{
        \begin{tabular}{lcccc}
            \toprule
            \textbf{Dataset} & \textbf{\# Dialogs} & \makecell{\textbf{Avg.} \\ \textbf{\# Turns}} & \makecell{\textbf{Avg. Length} \\ \textbf{(by token)}} & \makecell{\textbf{Avg. Utt. Length}\\ \textbf{(by token)}}  \\
            \midrule
            MSDialog   & \phantom{00}2,199 & 4.56 & \phantom{0}297.12 & \phantom{0}65.16     \\ 
            MANtIS   & \phantom{00}1,356 & 5.90 & \phantom{0}492.65 & 131.48     \\ %
            SOLISpeak \textbf{(Ours)}  & 316,697 & 4.57 & 1064.70 & 236.60   \\ 
            SOLITurbo (\textbf{Ours})   & 316,697    & 4.57          & 1097.10   & 241.80     \\       
            \bottomrule
        \end{tabular}
    }
    \caption{Statistics of intent-aware information-seeking dialog datasets. `T', `Dial', `Utt', and `L' refer to Turns, Dialogs, Utterances, and Length respectively.} %
    \label{tab:stat}
\end{table}

\subsection{SOLID-RL}
SOLID employs an utterance-by-utterance generation process. While this approach results in highly coherent dialogs that closely follow the dialog seed, it brings significant computational costs. To address this, we propose SOLID-RL, a method that enables the generation of an entire dialog in a single step. 
We leverage Direct Preference Optimization (DPO)~\citep{Rafailov2023-mv} for our RL-training. DPO requires chosen and rejected samples for training. We label the generated dialog in an utterance-by-utterance manner by SOLID as `chosen’ and dialogs generated by a single instruction as `rejected’. The prompt for generating a dialog given a single instruction is presented in Figure \ref{fig:gen_rejected_dialog} in the appendix. Since the quality of synthetic data generated by SOLID can vary significantly, establishing a dialog-level quality metric that can be used to guide the training process of SOLID-RL is crucial. 

Through human assessment, we observed that shorter dialogs exhibit higher quality. We aimed to inject this observed signal during training by designing prefixes constrained to the length of the dialog. These prefixes are independent of the 'chosen’ or `rejected’ labels of the dialog and are only injected at the beginning of the dialog given its length. For the shortest dialogs, we add the prefix “excellent dialog.” During inference, we use the prefix “excellent dialog” for all of our generations, regardless of the length of the dialog that is going to be generated. This approach triggers the LLM to generate dialogs that are more similar to higher-quality dialogs than to lower-quality ones. Our method is inspired by OpenChat \cite{wang2024openchat} that proposes prefix-conditioned training, where different labeled data are annotated by more expert assessors. However, in our setup, both datasets are generated by the same LLM, but the length has a direct impact on the quality of the data.

We define this approach as \textbf{L}ength-based quality estimation for \textbf{M}ixed-\textbf{Q}uality training (\textbf{LMQ}), where dialog length serves as a quality indicator (See Appendix~\ref{sec:llmq_inclusion} and \ref{sec:llmq_exclusion} for examples of low-quality training instances). We propose a simple yet effective heuristic. Before each dialog, we add a textual prompt representing its assessed quality: ``excellent quality dialog:'' for 1-3 utterances, ``good quality dialog:'' for 4 utterances, ``average quality dialog:'' for 5-10 utterances, ``poor quality dialog:'' for 11-15 utterances, and ``very poor quality dialog:'' for 16-20 utterances. Below is an example of distinct initial prompt tokens for the highest quality:

\noindent
{\scriptsize\texttt{\textbf{[3 utterances]}} \ \ {\color{gray}\textit{\textbf{}}} \textit{Seed\texttt{<|assisstant|>}}  \textit{\textbf{Excellent quality dialog:}}} \\
{\scriptsize\texttt{\textbf{[9 utterances]}} \ \ {\color{gray}\textit{\textbf{}}} \textit{Seed\texttt{<|assisstant|>}}  \textit{\textbf{Poor quality dialog:}}} 

Since longer dialogs often contain coherence issues, topic drifts, and repetitions, dialog length serves as a useful quality proxy. During inference, we assume that SOLID-RL has learned to distinguish higher-quality data distributions, considering that our objective is to exclusively generate high-quality responses by prompting them with:

{\scriptsize\texttt{\textbf{[for any dialog]}} \ \ {\color{gray}\textit{\textbf{}}} \textit{Seed\texttt{<|assisstant|>}}  \textit{\textbf{Excellent quality dialog:}}}

\subsection{Dataset Creation}
Using SOLID to generate dialog data results in a 300k conversational dataset called \textbf{SOLISpeak}. Subsequently, we train SOLID-RL on SOLISpeak to generate dialogs more efficiently, resulting in a second dataset of 300k conversations called \textbf{SOLITurbo}. Both datasets are large-scale and intent-aware, with SOLITurbo benefiting from the enhanced data quality and efficiency provided by the fine-tuning and optimization process of SOLID-RL. Details for both datasets are presented in Table \ref{tab:stat}.

\section{Experimental Setup}
\par

\header{Training and Evaluation}
\label{subsec:scenarios-solispeaksoliturbo}
We demonstrate the utility of SOLISpeak and SOLITurbo by using them to train IP models and evaluating their performance on existing human-annotated datasets. Given the large scale of our generated datasets, a small subset of the data is sufficient for effectively training a model. Therefore, we explore some training scenarios in which our generated datasets can be used independently or jointly with human-generated data to improve the trained models.
We evaluate the synthetic data in two training scenarios:

\begin{itemize}[nosep,leftmargin=*]
\item \textbf{Scenario 1: Synthetic-Only Training.} In this scenario, the models are trained exclusively on our synthetically generated data. We can utilize the entire dataset or strategically selected subsets, as using subsets reduces computational costs and potentially improves model effectiveness by focusing on the most relevant instances. To create these subsets, we employ a retrieval-based strategy: we randomly pick instances from the existing real dialog datasets, and then retrieve the top-k most relevant dialogs from SOLISpeak and SOLITurbo based on a dot product metric.
In other words, we treat the real dialogs as queries, and the synthetic dialogs as a document collection. Note that in this scenario, the actual dialogs are used for retrieval purposes only, and not for model training. We use two retrieval models: \textit{BM25}~\cite{robertson2009probabilistic,elastic-bm25} and a \textit{BERT-based bi-encoder}~\cite{transformers-mpnet}. We use 1,760 dialogs from the training partition of MSDialog-intent as queries, and retrieve $k=15$ dialogs from SOLISpeak/SOLITurbo for each, yielding 26,400 dialogs. After removing duplicates, we end up with 25,779 dialogs.\footnote{We experimented multiple values of $k$ $\in$ $[5, 20]$ in increments of $5$, and found $k=5$ as optimal, although the effectiveness across values was relatively similar.}

\item \textbf{Scenario 2: Mixed Real-Synthetic Training.} In this scenario, we train a model using a combination of human-labeled dialogs and synthetically generated data. We employ three sampling strategies to create balanced training sets. As in the previous paragraph, sampling is conducted on the synthetic data relative to the human-labeled data, to address label imbalance in the latter. \textit{SeqInt-Bal} samples from synthetic data (SOLISpeak/SOLITurbo) to ensure that, when combined with the human-labeled dialogs, the final dataset's intent sequence labels are balanced. SeqInt-Bal ensures that all intent sequences have at least 1000 data samples.
\textit{Int-Bal} works similarly to SeqInt-Bal, but focuses on balancing the individual intents rather than full intent sequences, ensuring that all intents are represented by the same number of samples.
Finally, to demonstrate the effectiveness of balanced sampling, we compare the results with \textit{Random-EQ} that randomly samples a fixed number of data points, ensuring a one-to-one correspondence.

\end{itemize}

\header{LLM for generating data} We chose Zephyr-7b-beta as our LLM for data generation, as it was one of the most competitive and robust 7B models at the time of our experiments.
\par
\header{LLM supervised fine-tuning} We use Huggingface's trainers for Supervised Fine-Tuning (SFT) of SOLID-RL. 
We use QLora 4bit~\citep{dettmers2023qlora} for efficient fine-tuning. We employ a batch size and gradient accumulation of 4, and train for 4000 steps for both SFT and DPO, which takes about 7 hours to be trained on a single NVIDIA GPU P6000 40GB GPU memory. %
\par
\header{Datasets} We use the MSDialog-Intent~\cite{InforSeek_User_Intent_Pred} dataset to extract real-world user intents, ensuring a realistic order and distribution. We also use it for evaluating the IP models trained on our synthetic dataset, along with MANtIS \cite{penha2019mantis}.

\header{Intent Prediction models} For intent prediction, we use the IP models:
\begin{enumerate*}[label=(\roman*)]
    \item \texttt{cnn-context-rep}, a top-performing model on MSDialog;
    \item BERT, which has shown state-of-the-art results on MANtIS;
    \item We also fine-tune T5 for intent prediction, adapting it for intent generation;
    \item \texttt{zephyr-7b-beta} in a few-shot intent prediction setup.
\end{enumerate*}
Further details and implementations link can be found in Appendix \ref{app:IP_models}.
\par
\header{Performance Assessment} We evaluate the effectiveness of our framework using MSDialog-intent dataset. We also assess our RL-trained LLM using the benchmark datasets from the Huggingface leaderboard \cite{huggingface_leaderboard}.
Although this leaderboard does not directly measure the conversational quality of chat models, it provides a useful signal to validate whether fine-tuning has introduced regressions on the base model's reasoning and truthfulness capabilities. Further details can be found in Appendix \ref{sec:solid_rl_llmbenchmark}. Moreover, we verified there is no data leakage between Zephyr and Mistral, and our chosen evaluation datasets. More details can be found in Appendix \ref{app:data_leakage}.

\begin{table}[t]
\centering
\caption{
The evaluation results of the `cnn-context-rep' model~\cite{qu2018MSDialog} on the MSDialog-intent dataset with utilizing all designed sampling methods for each scenario. We also trained on MANtIS as reference. The terms `\# of SOLISpeak' and `\# of MS' denote the number of dialogs sampled from our synthetically generated dataset (SOLISpeak) and the human-annotated MSDialog-intent dataset, respectively. `P' and `F1' refer to Precision and F1 score.}%
\label{tab:sampling_analysis}
\scalebox{0.76}{
 \begin{tabular}{ll|l|l|ccc}
     \toprule
     & \multirow{2}{*}{Sampling} & \multirow{2}{*}{\# of } & \multirow{2}{*}{\# of}  & \multirow{2}{*}{P} & \multicolumn{2}{c}{F1}  \\ \cmidrule{6-7}
     & method & SOLISpeak & MS & & Micro & Macro  \\ 
     \midrule
     \multicolumn{7}{l}{\textbf{Cross-Dataset Evaluation}} \\ 
     & MANtIS & 0 & 0 & .5324 & .5523 & .4871 \\ \midrule
     \multicolumn{7}{l}{\textbf{Scenario 1}: Synthetic-Only Training} \\  %
     & --- & 316,697 & 0 & .7167 & .7197 & .6406 \\
     & Bi-Encoder &  25,779 & 0 &  \underline{.7626} & \underline{.7989} & \underline{.7228} \\ %
     & BM25 & 25,779 & 0 & .7600 & .7745 & .7198 \\
     \midrule
     \multicolumn{7}{l}{\textbf{Scenario 2}: Mixed Real-Synthetic Training} \\ %
     & --- & 0 & 1,760 & .7321 & .7562 & .6713 \\ %
     & SeqInt-Bal & 111,889  & 1,760 & \textbf{.7641} &\textbf{.8076} & \textbf{.7246} \\ %
     & Int-Bal & 19,635 & 1,760 &  .7613 & .7857 & .7211 \\ %
     & Random-Eq & 1,760 & 1,760 & .7402 & .7640 & .6940 \\ %
     
     \bottomrule
 \end{tabular}
}
\end{table}

\begin{table*}[ht]
\centering
\caption{The Results of intent predictors on the most effective sampling method (SeqInt-Bal) and the most simple while effective (RandomEq) in addition to the using only human as training data. `P' refers to the precision.}
\label{tab:main_res}
\small{
\scalebox{0.80}{
    \begin{tabular}{l|l|ccc|ccc} 
    \toprule
    \multirow{2}{*}{Model}           & \multirow{2}{*}{Sampling method} &  \multicolumn{3}{c}{MSDialog}         & \multicolumn{3}{|c}{MANtIS}        \\  %
                    &              & P     &  \multicolumn{1}{c}{F1-Micro} &\multicolumn{1}{c}{F1-Macro}       & \multicolumn{1}{|c}{P}      & \multicolumn{1}{c}{F1-Micro} & \multicolumn{1}{c}{F1-Macro}      \\  \midrule %
    \multirow{3}{*}{cnn-context-rep} & SeqInt-Bal           & .764                 & .808                 & .725                 & .801                   & .786                 & .625                  \\
                                     & Random-Eq            & .740                 & .764                 & .694                 & .784                   & .744                 & .595                  \\
                                     & Only human           & .732                 & .756                 & .671                 & .777                   & .736                 & .572                  \\ 
    \midrule
    \multirow{3}{*}{BERT-base}            & SeqInt-Bal           & .810                 & .831                 & .743                 & .855                   & .814                 & .644                  \\
                                     & Random-Eq            & .762                 & .792                 & .711                 & .814                   & .772                 & .617                  \\
                                     & Only human           & .734                 & .758                 & .674                 & .779                   & .738                 & .591                  \\ 
    \midrule
    \multirow{3}{*}{T5-base}              & SeqInt-Bal           & .839                 & .868                 & .771                 & .877                   & .833                 & .669                  \\
                                     & Random-Eq            & .791                 & .828                 & .737                 & .824                   & .812                 & .650                  \\
                                     & Only human           & .750                 & .772                 & .685                 & .780                   & .745                 & .587                  \\  \midrule
    \multirow{2}{*}{T5-base on SOLID-RL} &  SeqInt-Bal & \textbf{.846} & \textbf{.879} & \textbf{.786} & \textbf{.881}  & \textbf{.839} & \textbf{.677}  \\ 
     &  Random-Eq  & .807 & .835 & .746 & .829  & .814 & .656 \\ \midrule
    Zephyr-fewshot  & 3-shots      & .416 & .479 & .405 & .486  & .437 & .344  \\ 
    \bottomrule
    \end{tabular}
}
}
\end{table*}
\section{Results}\label{sec:results}
We address the following research questions:
\begin{itemize}[nosep,leftmargin=*]
     \item \textbf{(RQ1)} How effective is SOLISpeak in training intent predictors across different scenarios with either zero or partial availability of human data, and to what extent does using the SOLITurbo dataset enhance the effectiveness of these intent predictors compared to SOLISpeak?
    \item \textbf{(RQ2)} What is the impact of self-seeding on SOLID compared to alternative external-seeding methods?
    \item \textbf{(RQ3)} How does SOLID-RL's ability to generate a full intent-aware dialog from a single zero-shot instruction improve computational efficiency compared to SOLID's utterance-by-utterance dialog generation?
\end{itemize}

\header{Main results (RQ1)} 
As a preliminary quality evaluation of the generated dialogs by SOLID, we manually reviewed a sample of 100 dialogs to ensure alignment between intents and utterances. Furthermore, table~\ref{tab:sampling_analysis} presents the impact of different sampling methods within each scenario on MSDialog-intent. We use the state-of-the-art IP method for this dataset, aiming to compare the utility of generated data across different sampling approaches and offer insights rather than evaluate various IP models. Our findings indicate that the most effective sampling method is sequence of intents balancing (SeqInt-Bal), which incorporates a higher proportion of SOLID data. Intent balancing is the second-best method, using approximately six times less SOLISpeak data. We find that for all sampling methods, training the IP model on a combination of human and SOLISpeak leads to higher effectiveness than using human data alone. Furthermore, we find that using a bi-encoder for sampling ranks as the second most effective approach. An interesting observation is that Random-Eq outperforms scenarios where only the human-labeled data or only SOLID-generated data are used. Training exclusively on SOLISpeak achieves slightly lower effectiveness than using human data, demonstrating SOLISpeak’s strong utility in cases where human data is unavailable or restricted for IP model training.
Moreover, we observe that the IP model trained solely on our dataset (Scenario 1) achieves a precision of 0.7167, compared to 0.5324 for a model trained on the MANtIS human-annotated dataset. This shows that training on our large and diverse synthetic datasets can build effective IP models that are generalizable to other datasets.
\par
To further explore \textbf{RQ1}, we investigate utility of the SOLISpeak and SOLITurbo datasets on an additional dataset and across various IP models. Focusing on Scenario 2, where a small human-generated dataset is available, we apply the most effective sampling method, SeqInt-Bal, alongside Random-EQ. We also consider Zephyr in a few-shot setup to determine whether it can serve as a slower but effective IP method.
Our main observations from Table \ref{tab:main_res} are: (1) utilizing our synthetic dataset improves all three of the supervised IP methods compared only using human data by a large margin; this proves the usefulness of our dataset in downstream tasks; (2) SOLID-RL generates more valuable data for IP, as the T5 model trained on a combination of human and SOLID-RL data achieves the highest effectiveness compared to other IP methods; (3) Zephyr cannot predict the intents effectively in a few-shot setup, showing the challenge of prediction for LLMs, even though they can generate content following specific intents. 
This aligns with previous studies' findings~\cite{weiwei2023diqad,unlocking2023} and can be due to the fact that following instructions is a more natural task for LLMs than acting as an IP method in a few-shot setup.
\par
\begin{figure*}[ht]
    \centering
    \scalebox{0.45}{\includegraphics[]{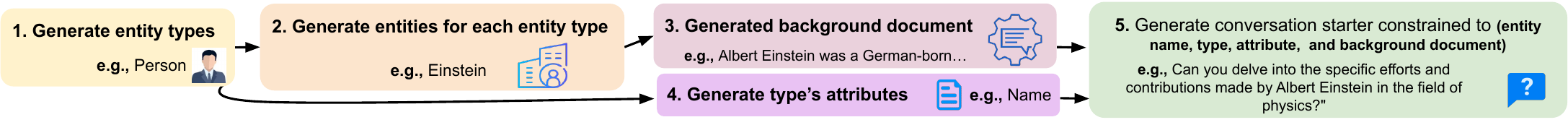}}
    \caption{The illustration of seed generation through self-seeding via SOLID.}
    \label{fig:other_llms_evaluation}
\end{figure*}

To better assess the necessity of constructing an intent-annotated dialog dataset for IP prediction, we evaluate Zephyr-7b and GPT-4o \cite{gpt4o} in a few-shot setting as intent predictors. Table \ref{fig:other_llms_evaluation} shows that intent prediction models trained on SOLID and SOLID-RL significantly outperform strong few-shot models such as GPT-4o, even with 20 provided examples. We would also like to highlight the efficiency and cost benefits of using the T5 model trained on SOLID or SOLID-RL data, which has only 220 million parameters. Deploying such a model is much more feasible compared to using larger LLMs with 70B or 405B parameters. As a result, our approach is not only more effective but also far more efficient in practical applications. While the T5-base (SOLID-RL) model performs significantly better than the other models, there are 'hard' dialogs on which all models fail. We present an example of such a dialog in Appendix \ref{app:hard_dialogs}.

\begin{table}[]
    \centering
    \scalebox{0.75}{
        \begin{tabular}{lccc}
            \toprule
            \textbf{Model} & \textbf{Precision} & \textbf{F1 (Micro)} & \textbf{F1 (Macro)} \\
            \midrule
            GPT-4o (20-shots)   & 0.546 & 0.589 & 0.424     \\ 
            Zephyr-7b (3-shots) & 0.416 & 0.479 & 0.405 \\    
            T5-base (SOLID-RL)  & 0.846 & 0.879	& 0.786 \\
            T5-base (SOLID)	& 0.839	& 0.868	& 0.771  \\       
            \bottomrule
        \end{tabular}
    }
    \caption{Evaluation of LLMs in few-shot settings as IP predictors.} %
    \label{tab:stat}
\end{table}
\par

\par
\header{Analyzing self-seeding impact (RQ2)} We examine the impact of self-seeding in comparison to three alternative external seeding methods. For external seeding, we employ various sampling approaches and prompt an LLM to generate 10,000 intent-aware dialogs about 1,000 entities. This is achieved by first generating ten different questions for each entity to serve as the starting point of the conversation. The alternative external-seeding methods are as follows: \textit{Top-most-popular entities}, we prompt the LLM with the top 1,000 most popular entities from Wikipedia; in \textit{Sampling from popular entities}, we prompt the LLM with 1,000 entities randomly sampled from the top 10,000 most popular entities on Wikipedia; and for \textit{Sampling random entities}, we randomly sample entities from Wikipedia. It is crucial to note that the sequence of intents remains consistent across this analysis. 
Table~\ref{tab:entityselection} shows that the IP models trained on data generated by the external seeding methods achieve lower effectiveness than those trained with self-seeding. This confirms the intuition behind self-seeding, which shows that it could improve the quality of data generated by the LLM.
\par
\header{SOLID-RL efficiency (RQ3)} We evaluate the efficiency of SOLID-RL, which can generate an entire dialog given a single instruction, compared to SOLID's utterance-by-utterance generation process that requires $N$ LLM passes to generate a dialog with $N$ turns. 
Using a batch size of 16, we measure the time required to generate 1,000 dialogs with each method. Our results indicate that SOLID takes 23,000 seconds to generate these dialogs, whereas SOLID-RL completes the task in just 2,000 seconds, making SOLID-RL nearly 12 times more efficient for intent-aware dialog generation. 
\par
\begin{table}
\centering
\caption{Comparative effectiveness of dialog generation using self-seeding vs.\ random selection of popular entities. The utterance classification model is T5 evaluated on MSDialog. `P' refers to the precision.}
\label{tab:entityselection}
\scalebox{0.80}{
    \begin{tabular}{lccc} 
    \toprule
                                & \multirow{2}{*}{P}        &  \multicolumn{2}{c}{F1} \\ \cmidrule{3-4}
                                &          & Micro & Macro  \\
    \midrule
    
Self-seeding \textbf{(ours)}          & .7085               & .7251               & .6240                 \\  \midrule
\multicolumn{4}{l}{\textbf{External-seeding methods}}    \\ 

Top-most-frequent entities & .6457               & .6604               & .5748                \\ 
Sampling frequent entities & .6460               & .6633               & .5769                \\ 
Sampling random entities & .5839               & .6101               & .4976                \\ 
    \bottomrule
    \end{tabular}
}
\end{table}
\begin{table}
\centering
\caption{Analyzing the impact of self-instructing and length-based quality estimation for mixed-quality training in SOLID-RL. `P' refers to the precision.}%
\label{tab:analyzing_solid_solidrl}
    \scalebox{0.85}{
        \begin{tabular}{lccc} 
        \toprule
        \multirow{2}{*}{Model}                            & \multirow{2}{*}{P}        &  \multicolumn{2}{c}{F1} \\ \cmidrule{3-4}
                                    &          & Micro & Macro  \\
        \midrule
        
    SOLID \textbf{(ours)}          & .708               & .725               & .624                 \\ 
    \,\,\,\,\textbf{W/o} self-instructing          & .515               & .537               & .428                 \\  \midrule %
    
    SOLID-RL \textbf{(ours)}          & .741               & .778               & .68                 \\ 
    \,\,\,\,\textbf{W/o} L-based MQ          & .532               & .554               & .457                 \\  
        \bottomrule
        \end{tabular}
    }
\end{table}
\section{Discussion}

\textbf{Multi-intent Self-instruction Analysis}. We assess the effect of multi-intent self-instructing on IP models by generating dialogs with 10,000 seeds using SOLID, and comparing the self-instruction approach to an alternative rule-based method that combines multiple instructions with “and.” The top section of Table \ref{tab:analyzing_solid_solidrl} reveals a significant performance difference between models trained on SOLID with self-instruction and those using the rule-based method, showing the potential of multi-intent self-instructing as a promising strategy.
\par
\header{Length-based mixed-quality training} Using the same 10,000 seeds as above, we examine the impact of LMQ on intent-aware dialog generation. To assess this, we train SOLID-RL both with and without LMQ’s class-conditioned input, starting from scratch in the latter case. For further details on the exclusion process, see Appendix \ref{sec:llmq_exclusion}. Results in the bottom section of Table \ref{tab:analyzing_solid_solidrl} reveal a significant advantage for SOLID-RL trained with LMQ, indicating that LMQ effectively optimizes and avoids low-quality dialogs impacting the LLM's performance. 
It also shows that conditioning the model with LMQ for high-quality output during inference aids in generating higher-quality data.
\par
\section{Conclusion}
This work presents a novel approach for automatically generating intent-aware, information-seeking dialog datasets through two zero-shot generation methods: SOLID and SOLID-RL. SOLID employs a novel zero-shot, multi-step process that uses self-seeding and self-instructing. SOLID-RL, optimized with supervised fine-tuning followed by DPO optimization, is trained on SOLID's output. With the ability to generate dialogs from a single instruction, SOLID-RL achieves an efficiency level 11 times greater than SOLID's utterance-by-utterance generation process. This efficiency is further enhanced by a unique length-based mixed-quality training method that dynamically assesses dialog quality. Leveraging these methods, we create two synthetic datasets that can significantly improve intent prediction accuracy. Despite LLMs' current limitations in intent prediction, our results highlight their potential to enhance training datasets. Our generated datasets provide valuable resources for other related tasks such as next intent prediction and conversational search. Our self-seeding and self-instructing approaches are adaptable to various conversation types and languages with minimal modifications.

\section*{Limitations}
In this paper, we demonstrated that LLMs can generate data to effectively train intent predictor models within the context of information-seeking dialogs. It is important to emphasize that this paper did not explore all aspects of evaluation, particularly the impact of biased information in the generated dialogs or the biases within the LLM's conversational abilities, that shows through factually incorrect generations. In our study, we did not examine how these biases affect the performance of intent predictors, which could lead to a tendency for certain utterances to be predicted with biased intents. Factually incorrect generations in the generated data (as a result of LLM's hallucination) are not likely to be harmful to the intent prediction models, as only information contained in human dialog will serve as input for the intent predictors in the test and real-world scenarios. Moreover, these predictors are limited to only the defined intents and cannot hallucinate new intents. 
We did not systematically investigate or quantify the impact of hallucinated data on the performance of intent predictor models and the potential bias they could introduce in the prediction. 
In the Appendix, we provide more information on this and estimate a generated entity that does not exist in sources like Wikipedia as a hallucination, but this is susceptible to errors.
While our method intuitively could work effectively with other large language models (LLMs), we have not empirically tested its performance across a broad range of LLMs, as we aimed to limit the scope of this paper. We leave the comprehensive evaluation of our method using different LLMs to future work, where potential variations in performance and biases can be thoroughly explored.
Our methodology can accommodate finer-grained taxonomies or additional annotations, thus its flexibility allows researchers to extend SOLID for more detailed tasks. Exploring these directions, including intent hierarchies and contextual annotations, is an exciting opportunity for future work.
\bibliography{custom}
\bibliographystyle{acl_natbib}
\newpage
\appendix
\section{Appendix}

\subsection{Intents}\label{app:intents}
In the context of an information-seeking conversation, we refer to intent as the role or function that an utterance plays in the conversation. It can include acts like requesting information, giving positive/negative feedback, requesting further details, or repeating the question. Table  \ref{table:intent_taxonomy} shows the list of 12 unique utterances that exist in the SOLID (ours) and MSDialog datasets. Intents can be used to help the chatbot understand the context and purpose of a user’s message and to determine how to respond appropriately. While similar in intuition, FQ (Follow-Up Question) and CQ (Clarifying Question) are distinct in their definitions, as established in prior studies. CQ focuses on clarification regarding specific aspects of a prior utterance, while FQ expands the discussion to related but broader topics. We followed these existing definitions to maintain consistency with established research and thus be able to compare with previous and future work.

\begin{table}[H]
    \centering
    \caption{Intent taxonomy and distribution in MSDialog}
    \setlength{\tabcolsep}{0pt} 
    \scriptsize
        \scalebox{0.90}{
            \begin{tabular}{llll}
                \toprule
                Code & Label & Description & \% \\ \midrule
                OQ & Original Question & The first question from the user to initiate the dialog. & 13 \\
                RQ & Repeat Question & Other users repeat a previous question. & 3 \\
                CQ & Clarifying Question & User or agent asks for clarifications. & 4 \\
                FD & Further Details & User or agent provides more details. & 14 \\
                FQ & Follow Up Question & User asks for follow up questions about relevant issues. & 5 \\
                IR & Information Request & Agent asks for information from users. & 6 \\
                PA & Potential Answer & A potential answer or solution provided by agents. & 22 \\
                PF & Positive Feedback & User provides positive feedback for working solutions. & 6 \\
                NF & Negative Feedback & User provides negative feedback for useless solutions. & 4 \\
                GG & Greetings/Gratitude & Greetings or expressing gratitude. & 22 \\
                JK & Junk & No useful information in the utterance. & 1 \\
                O & Others & Utterances cannot be categorized using other classes. & 1 \\ 
                \bottomrule
            \end{tabular}
        }
    \label{table:intent_taxonomy}
\end{table}

\subsection{Manually-crafted instructions for User and Agent Intents} \label{app:manual_instructions}
A total of 24 manually crafted instructions were created, covering 12 distinct intents, with separate instructions for both the user and agent sides, as shown in Tables \ref{tab:user_instructions} and \ref{tab:agent_instructions}. We want to note that the difference between the FQ and CQ is very fine, in linguistic nuance:

\begin{itemize}
    \item \textbf{Further Question (FQ)}: A question that builds on the previous information, i.e., it is a new question. This is emphasized by using the keyword \textbf{further}.  
    \underline{Example}: “How does this relate with the concept of supervised learning?” where \textbf{this} refers to the information in the previous turn.

    \item \textbf{Clarification Question (CQ)}: A question that asks for more rephrasing, i.e., clarification, of the previous turn, because the user/system cannot understand the meaning behind it.  
    \underline{Example}: “What exactly do you mean by image classification?” where the subject of the previous turn was still \textbf{image classification}.
\end{itemize}

The prompts were carefully engineered to capture this difference, and unfortunately, using the straight definition from Table \ref{table:intent_taxonomy} (which seems intuitive as human readers) did not yield satisfactory generations.

\begin{table}[ht]
    \centering
    \caption{Manually-crafted intent instructions. Actor type: Agent}
    \scalebox{0.8}{
        \begin{tabular}{l|p{8cm}}
            \toprule
            \textbf{Intent} & \textbf{Instruction} \\ \midrule
            CQ & Reply with one follow-up response in conversation style. \\ \midrule
            FD & Reply with further details in conversation style. \\ \midrule
            GG & Continue the conversation by expressing gratitude for the user's questions. \\ \midrule
            PA & Provide a potential solution or answer in conversation style. \\ \midrule
            IR & Ask the user to provide relevant information needed for their previous question. \\ \midrule
            OQ & Formulate an original question posed by an agent. \\ \midrule
            FQ & Formulate a follow-up question from an agent, seeking further clarification or information. \\ \midrule
            RQ & Now you are talking from the point of view of a third participant in the conversation. Repeat Question: \\ \midrule
            PF & Express satisfaction and appreciation for the conversation. \\ \midrule
            NF & Convey dissatisfaction for the previous response. \\ \midrule
            JK & Reply with gibberish information. It can contain emojis. \\ \midrule
            O & Reply with a system error. Return N/A \\ 
            \bottomrule
        \end{tabular}
    }
    \label{tab:agent_instructions}
\end{table}
\begin{table}[ht]
    \centering
    \caption{Manually-crafted intent instructions. Actor type: User}
    \scalebox{0.8}{
        \begin{tabular}{l|p{8cm}}
            \toprule
            \textbf{Intent} & \textbf{Instruction} \\ \midrule
            CQ & Reply with one question asking for clarification in conversation style. \\ \midrule
            FD & Reply with more details in conversation style. \\ \midrule
            GG & Continue the conversation by expressing gratitude for the agent's help. \\ \midrule
            PA & Provide a potential solution or answer in conversation style. \\ \midrule
            IR & Reply with relevant information. \\ \midrule
            OQ & Formulate the first question posed by a user that initiates a QA dialog. \\ \midrule
            FQ & Formulate a follow-up question from a user, seeking further clarification or information. \\ \midrule
            RQ & Now you are talking from the point of view of a third participant in the conversation. Repeat Question: \\ \midrule
            PF & Express satisfaction and appreciation for a working solution. \\  \midrule
            NF & Convey dissatisfaction for the previous response. \\  \midrule
            JK & Reply with gibberish information. It can contain emojis. \\  \midrule
            O & Reply with a system error. Return N/A \\ 
            \bottomrule
        \end{tabular}
    }
    \label{tab:user_instructions}
\end{table}

\subsection{Details of intent predictors}\label{app:IP_models}
We fine-tune BERT as the intent predictor model with a cross-entropy loss function. We fine-tune T5 for a seq-to-seq generation task with a cross-entropy loss function, using the prefix of "multilabel classification: " prompted to the input. For cnn-context-rep \cite{InforSeek_User_Intent_Pred}, we utilized the released code by the original paper and used the most effective reported hyperparameter. The implementation of the baselines are available at baseline folder of our repository which can be accessed at \href{https://anonymous.4open.science/r/solid\_repo/baselines/readme.md}{baselines folder}.

\subsection{Impact of hallucination} \label{app:hallucinations}
We categorize an entity as likely hallucinated if it cannot be located within Wikipedia. 
To facilitate this identification process, we leverage the Wikipedia search API~\cite{wikipeida-search-api}.
It is important to acknowledge that our methodology, being automated, would inevitably generate hallucinated content. 
In Table~\ref{tab:hallucination_analysis}, we examine the influence of hallucinated entities on subsequent tasks. Surprisingly, the results show that including hallucinated entities leads to higher effectiveness for intent prediction.
We believe the reason lies in the nature of the task itself: intent prediction focuses on understanding the intent of the text rather than ensuring the factually correctness of the text.

\begin{table}[H]
\centering
\caption{Analyzing impact of hallucination in SOLISpeak as training source on T5 intent predictor evaluated on MSDialog dataset.}
\label{tab:hallucination_analysis}
\scalebox{0.95}{
    \begin{tabular}{lccc} 
    \toprule
                                & \multicolumn{3}{c}{MSDialog}   \\ \cmidrule{2-4}
                                & P        &  \multicolumn{2}{c}{F1} \\ \cmidrule{3-4}
                                &          & Micro & Macro  \\
    \midrule
    Only hallucinated     & .7295 & .7498 & .6692           \\
    Only non-hallucinated & .7289 & .7502 & .6684           \\ 
    \midrule
    Full                  & .7321 & .7580 & .6710           \\
    \bottomrule
    \end{tabular}
}
\end{table}

\subsection{Self-seed generation process}\label{app:self_seeding}
\begin{figure*}[ht]
    \centering
    \scalebox{0.50}{\includegraphics[angle=90,origin=c]{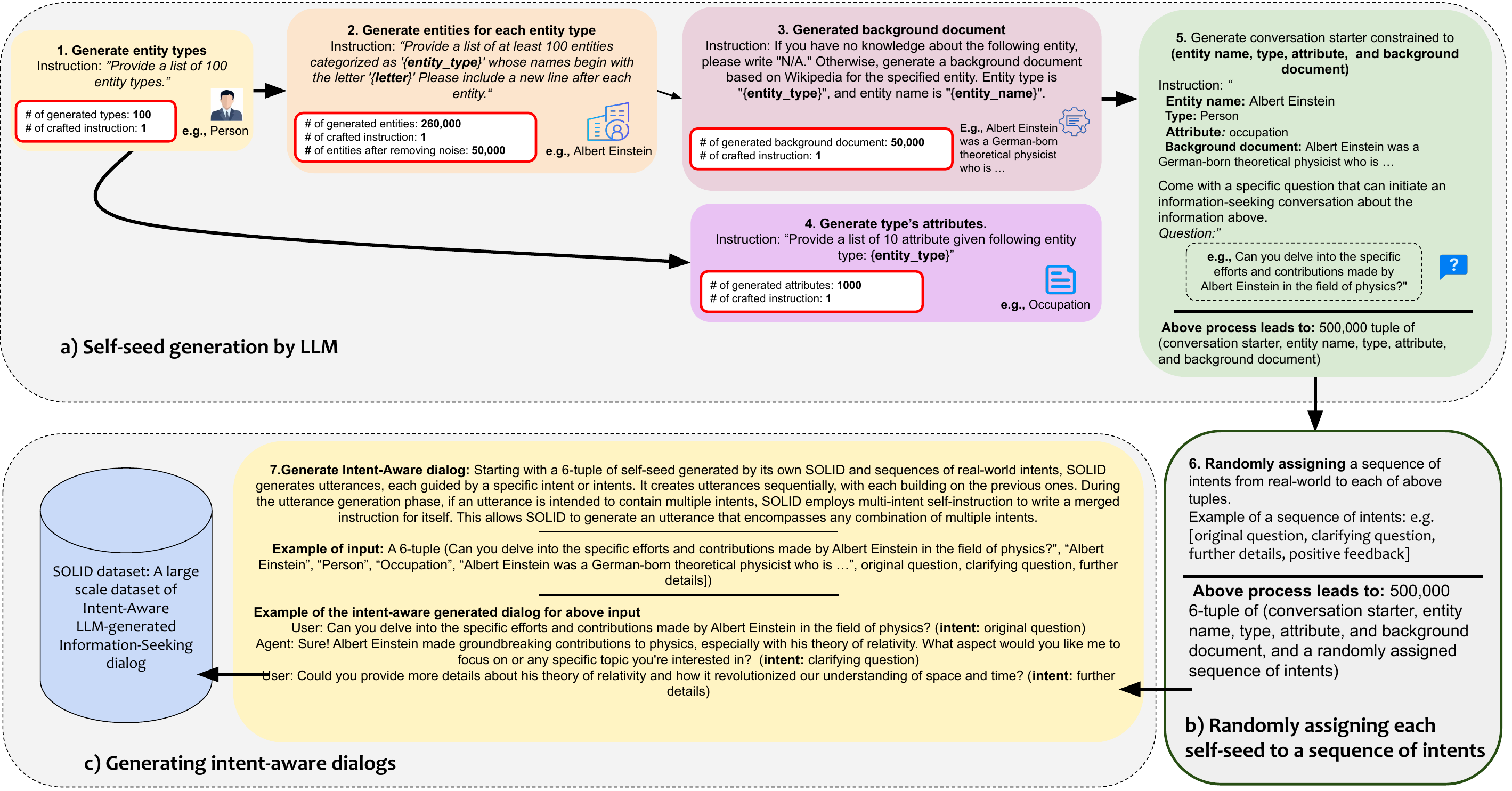}}
    \caption{Comprehensive depiction of the self-seed generation process in SOLID, illustrated through examples and prompts, alongside a summary of intent-aware dialog generation leveraging these self-seeds.}
    \label{fig:solid_selfseed_detailed}
\end{figure*}

Figure \ref{fig:solid_selfseed_detailed} presents a detailed illustration of the seed generation process. Additionally, Table \ref{tab:entity_types} provides all 98 distinct entity types we generated for SOLISpeak and SOLITurbo. Entity types that look like duplicates are removed.

\subsection{Prompts} \label{sec:prompts}
We have provided the content of all of the prompts used in our experiments. Our prompts are designed empirically, based on a comprehensive analysis of dialogue quality across various prompt configurations.
\subsubsection{Prompt for generating entity types}
We utilize a simple prompt asking `Provide a list of 100 entity types.' to generate entity types.

\subsubsection{Prompts for generating entity names}
\begin{figure}[H]
    \centering
    \includegraphics[width=0.9\linewidth]{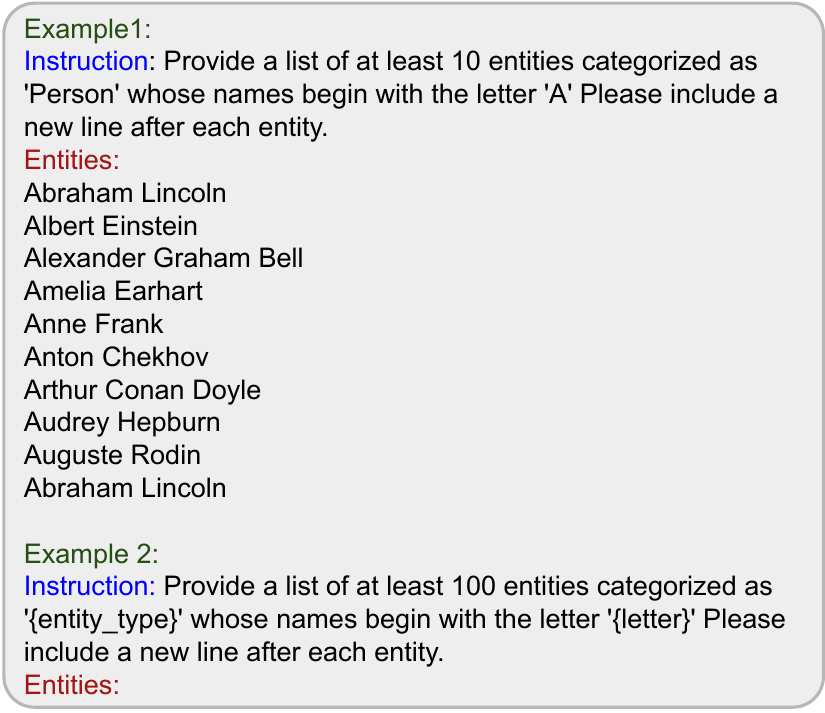}
    \caption{Prompt for generating entity names.}
    \label{fig:gen_entity_type}
\end{figure}

We empirically figured out that this task cannot be addressed effectively without providing a one-shot example to the LLM. Therefore, we provide 10 examples of generated entities, and ask the LLM to generate 100 entities. We exclude lines with more than 20 characters and remove special characters. The prompt is shown in Figure \ref{fig:gen_entity_type}.

\subsubsection{Prompt for generating conversation starters}
\begin{figure}[H]
    \centering
    \includegraphics[width=0.9\linewidth]{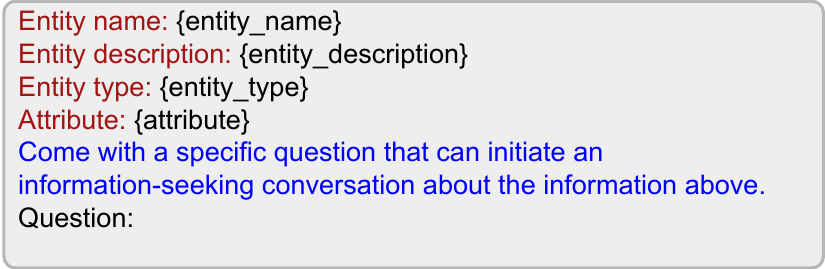}
    \caption{Prompt for generating a conversation starter}
    \label{fig:gen_conv_starter}
\end{figure}

The prompt for generating a conversation starter is shown in Figure \ref{fig:gen_conv_starter}.

\subsubsection{Prompt for building self-instruct instruction}
\begin{figure}[H]
    \centering
    \includegraphics[width=0.9\linewidth]{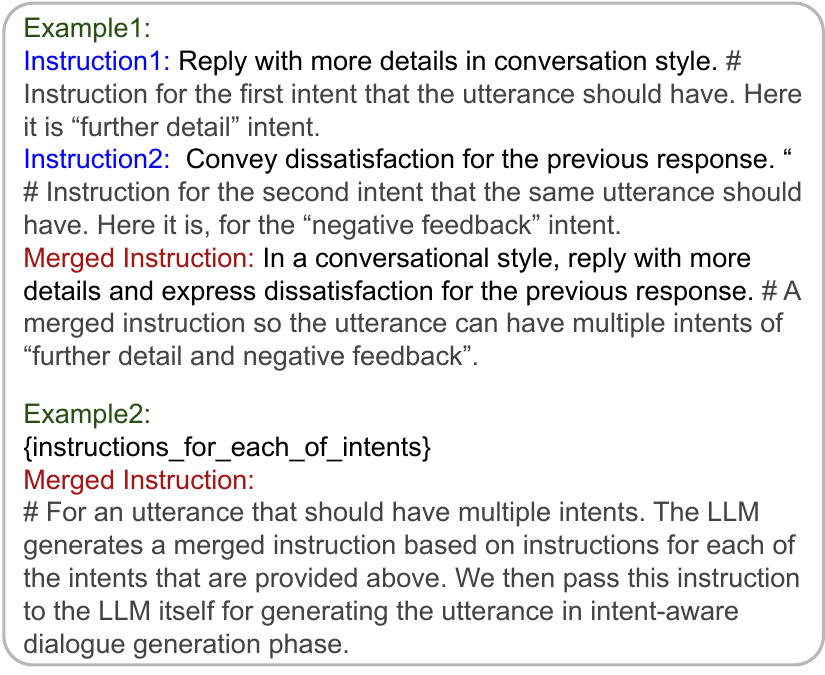}
    \caption{Prompt for generating self-instructions for utterances with multiple intents.}
    \label{fig:gen_self_instruct_combine_intents}
\end{figure}

When an utterance has more than one associated intent, we ask the LLM to generate it own instruction by merging the manually-crafted individual instructions presented in Tables \ref{tab:agent_instructions} and \ref{tab:user_instructions}. The prompt is shown in Figure \ref{fig:gen_self_instruct_combine_intents}.

\subsubsection{Prompt for generating an intent-aware conversation}
We combine the base prompt with the intent instructions. The base prompt is: `I will give you an entity, its type, a background document, a conversation history that ends in a...'. We empirically tested various prompts and refined our choices using ChatGPT, aiming to identify relatively the most effective prompt. The prompt we used was selected based on preliminary experiments that showed promising results in achieving the desired information-seeking dialog compared to other tested prompts. Using this prompt, we ask the LLM to generate the next utterance based on the current conversation history. Our current approach involves a one-way forward pass to generate the dialog sequentially. To further enhance data quality, we propose a backward pass—starting from the final utterance and working back to the first. In this reverse process, we would prompt the LLM to rewrite each utterance so that it better aligns with the intended next utterance, ensuring that the assigned intents are better reflected.

\begin{figure}[H]
    \centering
    \includegraphics[width=0.9\linewidth]{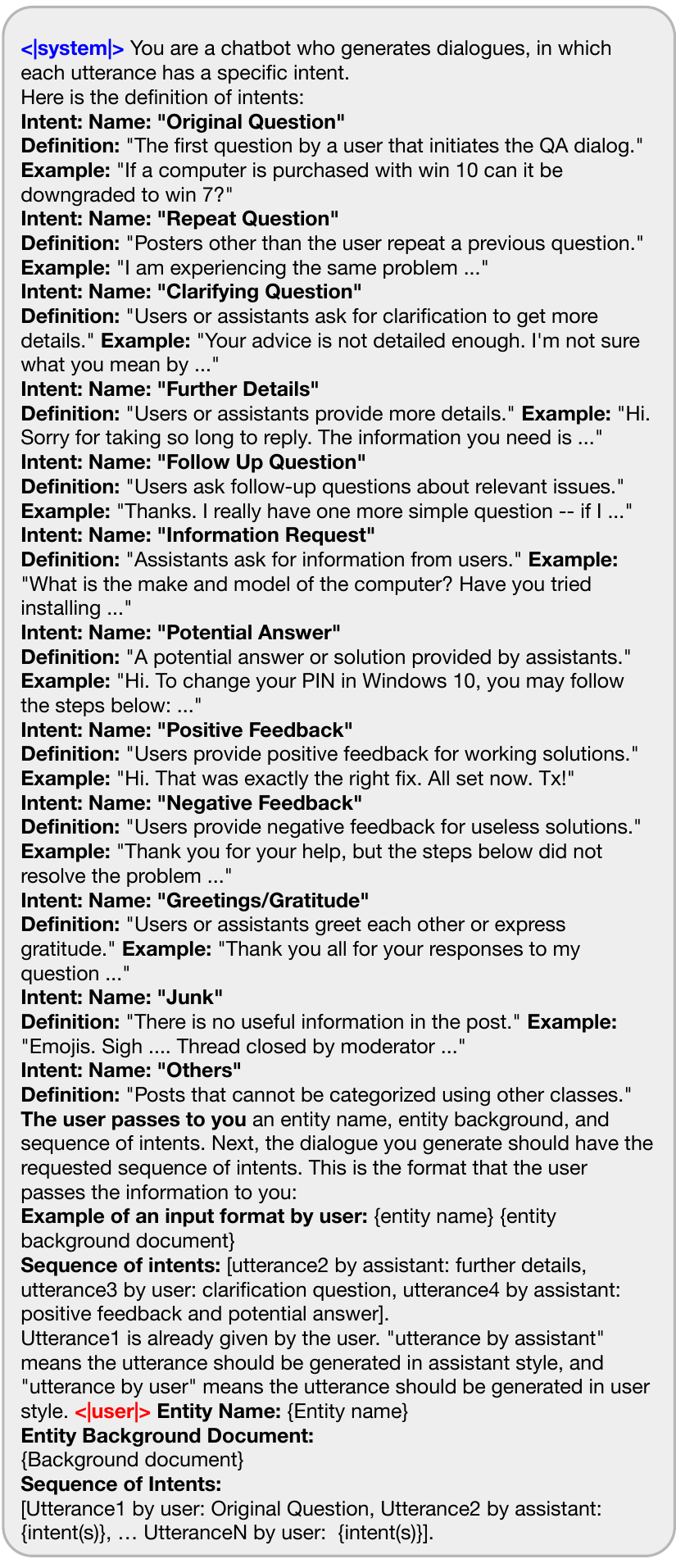}
    \caption{Prompt for generating `rejected' dialogs for DPO training.}
    \label{fig:gen_rejected_dialog}
\end{figure}

\subsection{Generating intent-aware dialogs.}
\label{sec:dialog_generation}
We first initialize the dialog with the conversation starter. For each intent \textit{i} in sequence \textbf{S}, we prompt the LLM to generate a new conversational turn. An intent can be either simple or complex. A simple intent is $i \in$ $keys(\textbf{I})$, whereas a complex intent is a combination of two or more intents $i = \{i_1, i_2, ... i_n \} \text{, where} \quad i_k \in \text{keys}(\textbf{I}) \text{ and} \quad k \sim \text{Uniform}(1, 12)$. If the intent is complex, we apply the \textit{multi-intent self-instructing} schema. For each intent in the sequence, the algorithm: (1) Retrieves its corresponding instruction or generates a merged instruction, (2) Concatenates the general instruction, entity, dialog history, and new intent-specific instruction, (3) Passes the message to the LLM to generate a new utterance, (4) Applies post-processing on the new utterance, and (5) Inserts the new utterance at the end of the dialog. When the whole sequence \textbf{S} has been processed, the algorithm returns the complete generated dialog \textbf{D}.

\subsection{Example of LMQ training instances.}
Whether a sample is selected or rejected during LMQ-based DPO training, a textual prefix on dialog length is added to construct a class-conditioned input. When LMQ is not used, this prefix is omitted. Examples for each setup are provided below.

\subsubsection{Including LMQ}\label{sec:llmq_inclusion}

We constrain the class-conditioned input only based on the dialog length. As a result, both the rejected and accepted samples have the same class-conditioned output "excellent quality dialog" due to their equal length in terms of the number of utterances.

\begin{figure}[H]
    \begin{mdframed}
    \textbf{excellent quality dialog:}
    \newline\newline
    Utterance 1:\newline
    Text: Can you delve into the specific efforts and contributions made by Albert Einstein in the field of physics?\newline
    Intent: original question
    \newline\newline
    Utterance 2:\newline
    Text: Sure! Albert Einstein made groundbreaking contributions to physics, especially with his theory of relativity. What aspect would you like me to focus on or any specific topic you're interested in?\newline
    Intent: clarifying question
    \newline\newline
    Utterance 3:\newline
    Text: Could you provide more details about his theory of relativity and how it revolutionized our understanding of space and time?\newline
    Intent: further details
    \end{mdframed}
    \caption{A chosen training sample with LMQ.}
\end{figure}

\begin{figure}[H]
    \begin{mdframed}
    \textbf{excellent quality dialog:}
    \newline\newline
    Utterance 1:\newline
    Text: Can you delve into the specific efforts and contributions made by Albert Einstein in the field of physics?\newline
    Intent: original question
    \newline\newline
    Utterance 2:\newline
    Text: Absolutely! Albert Einstein made revolutionary advancements in the field of physics, particularly through his theory of relativity.\newline
    Intent: clarifying question
    \newline\newline
    Utterance 3:\newline
    Text: How is the weather?\newline
    Intent: further details
    \end{mdframed}
    \caption{A rejected training sample with LMQ.}
\end{figure}

\subsubsection{Excluding LMQ}\label{sec:llmq_exclusion}

\begin{figure}[H]
    \begin{mdframed}
    Utterance 1:\newline
    Text: Can you delve into the specific efforts and contributions made by Albert Einstein in the field of physics?\newline
    Intent: original question
    \newline\newline
    Utterance 2:\newline
    Text: Sure! Albert Einstein made groundbreaking contributions to physics, especially with his theory of relativity. What aspect would you like me to focus on or any specific topic you're interested in?\newline
    Intent: clarifying question
    \newline\newline
    Utterance 3:\newline
    Text: Could you provide more details about his theory of relativity and how it revolutionized our understanding of space and time?\newline
    Intent: further details
    \end{mdframed}
    \caption{A chosen training sample without LMQ.}
\end{figure}

\begin{figure}[H]
    \begin{mdframed}
    Utterance 1:\newline
    Text: Can you delve into the specific efforts and contributions made by Albert Einstein in the field of physics?\newline
    Intent: original question
    \newline\newline
    Utterance 2:\newline
    Text: Absolutely! Albert Einstein made revolutionary advancements in the field of physics, particularly through his theory of relativity.\newline
    Intent: clarifying question
    \newline\newline
    Utterance 3:\newline
    Text: How is the weather?\newline
    Intent: further details
    \end{mdframed}
    \caption{A rejected training sample without LMQ.}
\end{figure}

\subsection{SOLID-RL on LLM benchmarks}
\label{sec:solid_rl_llmbenchmark}
\begin{figure*}
    \centering
    \includegraphics[width=1\linewidth]{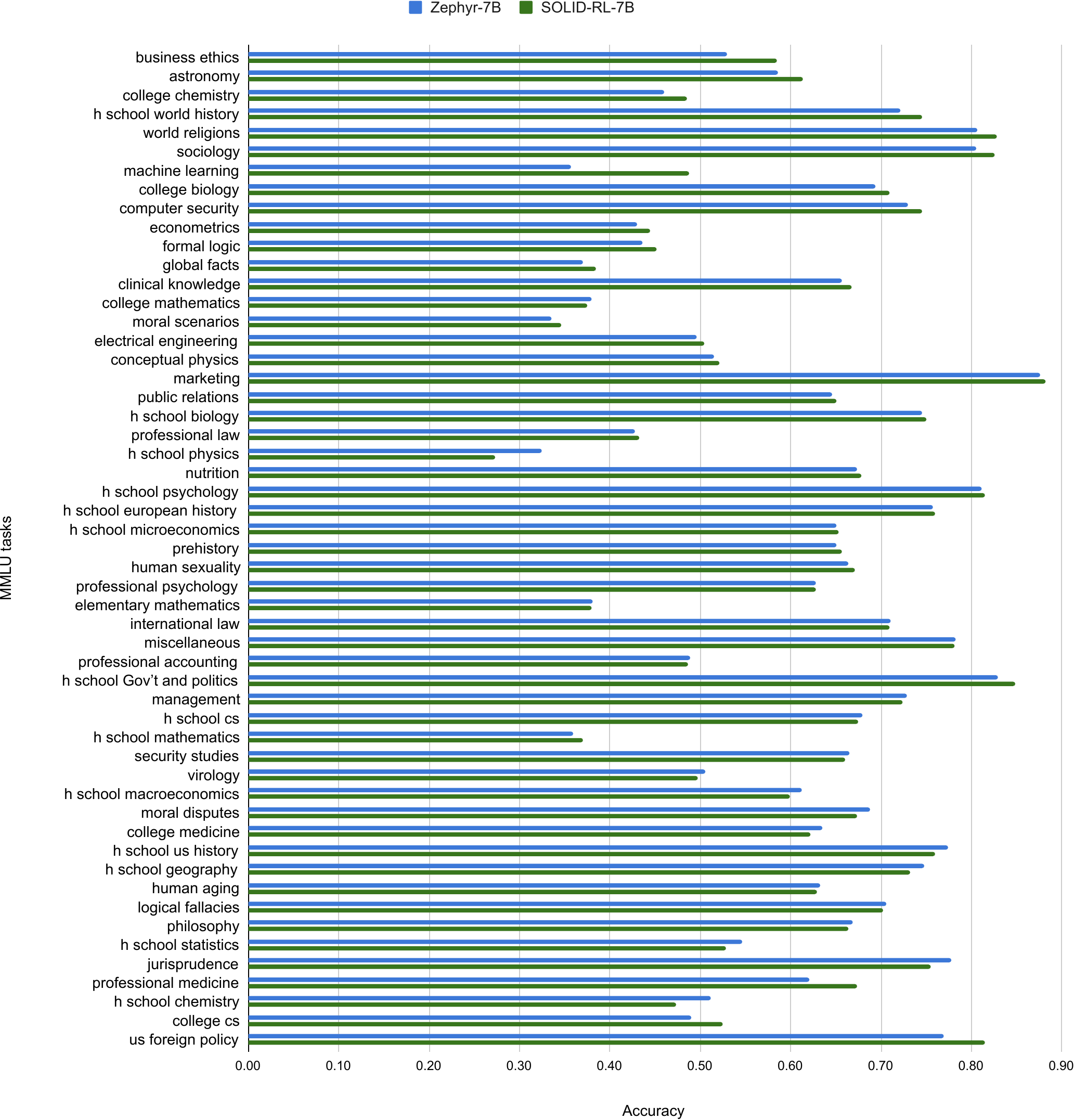}
    \caption{Effectiveness of SOLID-RL-7B vs. Zephyr-7B in terms of accuracy over tasks of Measuring Massive Multitask Language Understanding (MMLU) benchmark~\cite{hendrycks2020measuring}.}
    \label{fig:mmlu_tasks}
\end{figure*}

We analyze the effect of using SOLID-RL through performance of Zephyr-7b-beta on multiple LLM benchmarks on the Open LLM Leaderboard~\citep{huggingface_leaderboard}. We gain insights into whether SOLID-RL introduces regressions on the base model's reasoning and truthfulness capabilities. Table \ref{tab:llm_ability} provides an in-depth analysis of the impact of different training strategies for SOLID-RL on its performance. Our results demonstrate that SOLID-RL enhances Zephyr's performance on both MMLU~\citep{Hendrycks2021-tt} and Truthful QA~\citep{Lin2022-ms} benchmarks.
Our finding is relatively aligned with previous studies that show LLMs can enhance themselves when trained on high-quality self-generated data~\citep{huang2022large,yuan2024self,chen2024grath,chen2024self}.
The results indicate that excluding length-based mixed-quality training can significantly harm the general capability of SOLID-RL. We highlight the positive impact of DPO after SFT. We further investigate the effect of reversing chosen/rejected outputs in DPO training by using the output generated by SOLID as the rejected output during DPO optimization. This setup results in lower performance, confirming the chosen setup in SOLID-RL. We also experiment with training SOLID-RL alongside Zephyr's training dataset in a multi-task setup; we found that this led to a decrease in the effectiveness of the LLM and to a generation of lower-quality dialogs. Our analysis suggests that SOLID-RL is the most robust method compared to the variations investigated.
Furthermore, we demonstrate the performance of SOLID-RL on 57 tasks on MMLU compared with Zephyr-7b, as illustrated in Figure~\ref{fig:mmlu_tasks}. An example of an MMLU task is illustrated in Figure \ref{fig:example_mmlu}.

\begin{table*}[]
\centering
\caption{The evaluation results on the Open LLM Leaderboard~\citep{huggingface_leaderboard}, which measures the performance of LMs across four multiclass classification tasks: ARC~\citep{Clark2018-eq}, HellaSwag~\citep{Zellers2019-tj}, MMLU~\citep{Hendrycks2021-tt}, and Truthful QA~\citep{Lin2022-ms}. There are 57 tasks in MMLU - a detailed report on the effectiveness of SOLID-RL on all of them is shown in Figure \ref{fig:mmlu_tasks}. The bottom section of the table provides various analyses on SOLID-RL including: an ablation study on our length-oriented mixed-quality training, ablation study on DPO, reverting the preference data in DPO, and an analysis on multi-task training. The base model of SOLID-RL-7B is Zephyr-7B.}
\label{tab:llm_ability}
\scalebox{0.95}{
    \begin{tabular}{l|c|c|c|c} 
        \toprule
            Model                       & \multicolumn{4}{c}{Accuracy}        \\ 
                                   & Truthful QA & Hella Swag & MMLU  & ARC   \\ \midrule
        UltraLM-13B~\cite{touvron2023llama} & 52.00       & 61.32      & 50.45 & 57.25  \\
        Zephyr-7B~\cite{tunstall2023zephyr}   & 57.44       &\textbf{ 84.52}      & 61.44 & \textbf{62.03} \\ 
        SOLID-RL-7B \textbf{(Ours)}               & \textbf{58.13}       & \underline{83.82}      & \textbf{61.68}& \underline{60.78}   \\ \midrule
        \multicolumn{5}{l}{\textbf{Analysis on SOLID-RL-7B}:} \\
        \,\,\,W/o length-oriented MixedQ training & 55.25       & 80.69      & 58.40 & 59.24    \\
        \,\,\,W/o DPO  (i.e., only SFT)         & 53.77       & 81.72      & 60.47 & 59.73   \\
        \,\,\,Reverse chosen/rejected for DPO   & 52.36       & 79.82      & 60.14 & 58.96   \\ 
        \,\,\,W/ multi-task training        & 57.58       & 82.53      & 59.60 & 59.56  \\
        \bottomrule
    \end{tabular}
}
\end{table*}

\begin{algorithm*}[]
\caption{Generate Dialog}\label{algo:generate_dialog}
\small{
\textbf{Input:} $\mathbf{G}$ (global task instruction), $\mathbf{I}$ (intent instructions dictionary), $\mathbf{E}$ (entity), $\mathbf{D}_0$ (conversation starter), $\mathbf{S}$ (sequence of intents)\\
\textbf{Output:} $\mathbf{D}$ (generated multi-turn dialog)}
\begin{multicols}{2}
\small{
    \begin{algorithmic}[1]
    \State Initialize dialog $\mathbf{D} \leftarrow \mathbf{D}_0$
    \For{each intent $i$ in sequence $\mathbf{S}$}
        \If{\texttt{multiple\_intents($i$)}}
            \State Set $\mathbf{instruction} \leftarrow$ \texttt{SelfInstruct($i$)}
        \Else
            \State Retrieve $\mathbf{instruction} \leftarrow \mathbf{I}[s]$
        \EndIf
        \State $\mathbf{message} \leftarrow \texttt{Concat($\mathbf{G}, \mathbf{C}, \mathbf{D}, \mathbf{instruction}$)}$
        \State $\mathbf{utterance} \leftarrow$ \texttt{GenUtterance($\mathbf{message}$)}
        \State $\mathbf{utterance} \leftarrow$ \texttt{PostProcess($\mathbf{utterance}$)}
        \State $\mathbf{D} \leftarrow$ \texttt{InsertNewTurn($\mathbf{D}, \mathbf{utterance}$)}
    \EndFor
    \State \textbf{return} $\mathbf{D}$
    \end{algorithmic} }
\end{multicols}
\end{algorithm*}

\begin{figure}[H]
\centering
\fbox{\begin{minipage}{7 cm}
A 6-sided die is rolled 15 times and the results are: side 1 comes up 0 times; side 2: 1 time; side 3: 2 times; side 4: 3 times; side 5: 4 times; side 6: 5 times. Based on these results, what is the probability of side 3 coming up when using Add-1 Smoothing?\\
(A) 2/15 \quad \textbf{(B) 1/7} \quad (C) 3/16 \quad (D) 1/5
\end{minipage}}
\caption{A Machine Learning example of MMLU benchmark~\cite{hendrycks2020measuring}.}
\label{fig:example_mmlu}
\end{figure}

\subsection{Safeguard Data Leakage}\label{app:data_leakage}
To ensure there is no data leakage between the evaluation and training datasets used for Zephyr, we reviewed the Zephyr and Mistral training documentation~\cite{tunstall2023zephyr}. We checked if any of the BEIR collection data, such as MSDialog or MANtIS, exist in the training data of Mistral or Zephyr. We made sure that there is no mutual dataset within the used LLM training data, X~\cite{tunstall2023zephyr} and our evaluation datasets.

\subsection{Hard Examples}\label{app:hard_dialogs}

We include an example of a 'hard' dialog, which has been wrongly predicted by all models in our experimental setup:

\begin{quote}
\small
\textbf{User}: Hi, the best way to view photos in Windows 10 is by using the Photos app. For us to better assist you, may we know how you are accessing your photos?  

\textbf{Agent}: What changes have you made prior to getting this issue? What troubleshooting steps have you done so far? Can you provide a screenshot of the error message you're getting?

\textbf{User}: We'll be waiting for your response.
\end{quote}

For this specific example, the predicted intents by each model are:

\begin{itemize}
    \item \textbf{GPT-4o}: \texttt{['Further Details']}
    \item \textbf{T5 (SOLID)}: \texttt{['Potential Answer', 'Clarification Question']}
    \item \textbf{T5 (SOLID-RL)}: \texttt{['Potential Answer', 'Clarification Question']}
\end{itemize}

while the ground truth intents are:

\begin{itemize}
    \item \texttt{['Information Request', 'Potential Answer', 'Clarification Question']}
\end{itemize}

\noindent While it may seem clear to a human that the message encompasses all three intents in the ground truth, neither GPT-4o nor our models can successfully detect all of them. However, T5 models trained on SOLID and SOLID-RL are able to capture two out of the three ground truth intents, whereas GPT-4o fails to detect any of the correct intents.

\clearpage

\begin{table*}[]
  \centering
    \caption{Diverse set of 98 entity types generated via SOLID.}
    \label{tab:entity_types}
    \scalebox{0.60}{
    \begin{tabular}{llllll}
    \toprule
    \makebox[0.6cm][c]{\faUser{}} Person & \makebox[0.6cm][c]{\faHistory{}} Historical Event & \makebox[0.6cm][c]{\faCoffee{}} Beverage & \makebox[0.6cm][c]{\faMeteor{}} Galaxy & \makebox[0.6cm][c]{\faPenFancy{}} Author & \makebox[0.6cm][c]{\faVenus{}} Deity \\
      \makebox[0.6cm][c]{\faBook{}} Book & \makebox[0.6cm][c]{\faGalacticRepublic{}} Constellation & \makebox[0.6cm][c]{\faPaintBrush{}} Artist & \makebox[0.6cm][c]{\faHatWizard{}} Mythology & \makebox[0.6cm][c]{\faHeadphones{}} Song & \makebox[0.6cm][c]{\faSatellite{}} Astronomical Object \\
    \makebox[0.6cm][c]{\faUserTie{}} Politician &  \makebox[0.6cm][c]{\faDragon{}} Mythical Creature & \makebox[0.6cm][c]{\faPaw{}} Animal & \makebox[0.6cm][c]{\faBolt{}} Natural Disaster 
    & \makebox[0.6cm][c]{\faTheaterMasks{}} Actor & \makebox[0.6cm][c]{\faGhost{}} Supernatural Being \\
    \makebox[0.6cm][c]{\faGamepad{}} Game & \makebox[0.6cm][c]{\faCloudSunRain{}} Weather Phenomenon & \makebox[0.6cm][c]{\faRecordVinyl{}} Musician & \makebox[0.6cm][c]{\faMask{}} Character (Fictional) & \makebox[0.6cm][c]{\faTablet{}} Gadget & \makebox[0.6cm][c]{\faVirus{}} Disease \\
    \makebox[0.6cm][c]{\faCameraRetro{}} Celebrity & \makebox[0.6cm][c]{\faBuilding{}} Organization & \makebox[0.6cm][c]{\faLaptopCode{}} Software & \makebox[0.6cm][c]{\faCapsules{}} Medication & \makebox[0.6cm][c]{\faFilm{}} Actress & \makebox[0.6cm][c]{\faIndustry{}} Company \\
    \makebox[0.6cm][c]{\faAndroid{}} App & \makebox[0.6cm][c]{\faSyringe{}} Medical Procedure & \makebox[0.6cm][c]{\faRunning{}} Athlete & \makebox[0.6cm][c]{\faFolderOpen{}} Government Agency & \makebox[0.6cm][c]{\faLink{}} Website & \makebox[0.6cm][c]{\faBalanceScale{}} Law \\
    \makebox[0.6cm][c]{\faLandmark{}{}} Historical Figure & \makebox[0.6cm][c]{\faHeart{}} Nonprofit Organization & \makebox[0.6cm][c]{\faHashtag{}} Social Media Platform & \makebox[0.6cm][c]{\faGavel{}} Legal Case & \makebox[0.6cm][c]{\faChartLine{}} Entrepreneur & \makebox[0.6cm][c]{\faHospital{}} Hospital  \\
    \makebox[0.6cm][c]{\faMobile{}} Device & \makebox[0.6cm][c]{\faBrain{}} Political Ideology & \makebox[0.6cm][c]{\faLightbulb{}} Inventor & \makebox[0.6cm][c]{\faUniversity{}} University & \makebox[0.6cm][c]{\faGuitar{}} Instrument & \makebox[0.6cm][c]{\faFistRaised{}} Social Movement \\
    \makebox[0.6cm][c]{\faFlask{}} Scientist & \makebox[0.6cm][c]{\faCalendarDay{}} Event & \makebox[0.6cm][c]{\faWrench{}} Tool & \makebox[0.6cm][c]{\faBookOpen{}} Philosophy & \makebox[0.6cm][c]{\faDivide{}} Mathematician & \makebox[0.6cm][c]{\faSchool{}} School \\
    \makebox[0.6cm][c]{\faCouch{}} Furniture & \makebox[0.6cm][c]{\faChurch{}} Religion & \makebox[0.6cm][c]{\faScroll{}} Philosopher & \makebox[0.6cm][c]{\faMapSigns{}} Place & \makebox[0.6cm][c]{\faTshirt{}} Clothing & \makebox[0.6cm][c]{\faTree{}} Folklore \\
    \makebox[0.6cm][c]{\faCompass{}} Explorer & \makebox[0.6cm][c]{\faFlag{}} Country & \makebox[0.6cm][c]{\faPalette{}} Artwork & \makebox[0.6cm][c]{\faLanguage{}} Language  & \makebox[0.6cm][c]{\faFeather{}} Poet & \makebox[0.6cm][c]{\faCity{}} City \\
    \makebox[0.6cm][c]{\faPaintRoller{}} Painting & \makebox[0.6cm][c]{\faTrademark{}} Brand & \makebox[0.6cm][c]{\faCameraRetro{}} Photographer & \makebox[0.6cm][c]{\faArchway{}} Architectural Structure & \makebox[0.6cm][c]{\faHammer{}} Sculpture & \makebox[0.6cm][c]{\faTshirt{}} Fashion Brand \\
    \makebox[0.6cm][c]{\faNewspaper{}} Journalist & \makebox[0.6cm][c]{\faDrumSteelpan{}} Festival & \makebox[0.6cm][c]{\faCheese{}} Cuisine & \makebox[0.6cm][c]{\faShoppingBag{}} Style (Fashion, Art) &  \makebox[0.6cm][c]{\faBullhorn{}} Activist & \makebox[0.6cm][c]{\faComments{}} Conference \\
    \makebox[0.6cm][c]{\faCarrot{}} Recipe & \makebox[0.6cm][c]{\faTheaterMasks{}} Genre (Music, Film) & \makebox[0.6cm][c]{\faFutbol{}} Sportsperson & \makebox[0.6cm][c]{\faConciergeBell{}} Service & \makebox[0.6cm][c]{\faVial{}} Scientific Concept & \makebox[0.6cm][c]{\faMicrochip{}} Technology \\
    \makebox[0.6cm][c]{\faFootballBall{}} Sport & \makebox[0.6cm][c]{\faBox{}} Product & \makebox[0.6cm][c]{\faAtom{}} Chemical Element & \makebox[0.6cm][c]{\faCode{}} Programming Language & \makebox[0.6cm][c]{\faUsers{}} Team & \makebox[0.6cm][c]{\faUtensils{}} Food \\
    \makebox[0.6cm][c]{\faDotCircle{}} Particle & \makebox[0.6cm][c]{\faVideo{}} Movie 
    & \makebox[0.6cm][c]{\faTrophy{}} League & \makebox[0.6cm][c]{\faCar{}} Vehicle & \makebox[0.6cm][c]{\faGlobe{}} Planet & \makebox[0.6cm][c]{\faHourglassHalf{}} Historical Period \\
    \makebox[0.6cm][c]{\faSeedling{}} Plant & \makebox[0.6cm][c]{\faStar{}} Star & \\
    \bottomrule
  \end{tabular}}
\end{table*}

\newpage

\clearpage

\end{document}